%% file: ms.tex
\documentclass[runningheads]{llncs}
\usepackage{graphicx}
\usepackage{comment}
\usepackage{amsmath,amssymb} 
\usepackage{color}

\usepackage{subcaption}
\usepackage{amsmath}
\usepackage{amssymb}
\usepackage{bbm}

\usepackage{algorithm}
\usepackage{algpseudocode}
\algrenewcommand\algorithmiccomment[2][\scriptsize]{{#1\hfill\(\triangleright\) #2}}

\input{math_commands.tex}

\usepackage{xspace}
\makeatletter
\DeclareRobustCommand\onedot{\futurelet\@let@token\@onedot}
\def\@onedot{\ifx\@let@token.\else.\null\fi\xspace}

\def\eg{\emph{e.g}\onedot}

\def\etc{\emph{etc}\onedot} 
 
\def\etal{\emph{et al}\onedot}
\makeatother

\usepackage{array}
\usepackage{bbding}
\usepackage{pifont}
\usepackage{wasysym}
\usepackage{amssymb}
\usepackage{multicol}
\newcolumntype{P}[1]{>{\centering\arraybackslash}p{#1}}
\begin{document}

\pagestyle{headings}
\mainmatter

\title{Dynamically Throttleable Neural Networks (TNN)} 


\titlerunning{Dynamically Throttleable Neural Networks (TNN)}

\author{Hengyue Liu\inst{1,2},
Samyak Parajuli\inst{1,3},
Jesse Hostetler\inst{4},
Sek Chai\inst{1},
Bir Bhanu\inst{2}
}
\authorrunning{Liu, et al.}
\institute{Latent AI, Princeton, NJ \and
UC Riverside, Riverside, CA \and
UC Berkeley, Berkeley, CA \and
SRI International, Princeton, NJ}

\maketitle

\input{0_Abstract}
\input{1_Intro}
\input{2_RelatedWork}
\input{3_TechApproach}

\input{4_Experiments}

\input{5_Analysis}

\input{6_Conclusion}

%
%
\bibliographystyle{splncs04}
\bibliography{egbib}
\end{document}

%% file: math_commands.tex
\usepackage{amsmath,amsfonts,bm}

\newcommand{\Defined}{\stackrel{\textrm{\scriptsize def}}{=}}

\def\eqref#1{equation~\ref{#1}}

\def\1{\bm{1}}

\newcommand{\test}{\mathcal{D_{\mathrm{test}}}}

\def\va{{\bm{a}}}

\def\vf{{\bm{f}}}
\def\vg{{\bm{g}}}

\def\vp{{\bm{p}}}

\def\vw{{\bm{w}}}

\def\evg{{g}}

\DeclareMathAlphabet{\mathsfit}{\encodingdefault}{\sfdefault}{m}{sl}
\SetMathAlphabet{\mathsfit}{bold}{\encodingdefault}{\sfdefault}{bx}{n}

\newcommand{\E}{\mathbb{E}}



%% file: 0_Abstract.tex
\begin{abstract}

Conditional computation for Deep Neural Networks (DNNs) reduce overall computational load and improve model accuracy by running a subset of the network. In this work, we present a runtime throttleable neural network (TNN) that can adaptively self-regulate its own performance target and computing resources. We designed TNN with several properties that enable more flexibility for dynamic execution based on runtime context. TNNs are defined as throttleable modules gated with a separately trained controller that generates a single utilization control parameter. We validate our proposal on a number of experiments, including Convolution Neural Networks (CNNs such as VGG, ResNet, ResNeXt, DenseNet) using CiFAR-10 and ImageNet dataset, for object classification and recognition tasks. We also demonstrate the effectiveness of dynamic TNN execution on a 3D Convolustion Network (C3D) for a hand gesture task. Results show that TNN can maintain peak accuracy performance compared to vanilla solutions, while providing a graceful reduction in computational requirement, down to 74\% reduction in latency and 52\% energy savings.

\end{abstract}

%% file: 1_Intro.tex
\section{Introduction}

Deep learning models are typically trained offline to produce models with a static allocation of compute and memory resource. However, the conditions in real-world setting are often different, whereby the runtime inference is neither optimal from an accuracy or efficiency perspective. The problem lies in the current training approaches that produce static models that occupy a single point in the trade-space between performance and resource use. This paper presents an approach to creating throttleable neural networks (TNNs) system. We use the term ''throttleable'' to refer to the ability of the model to adaptively balance performance and resource use in response to a control signal. 

\begin{figure*}[t]
\centering
\includegraphics[width=1.0\textwidth]{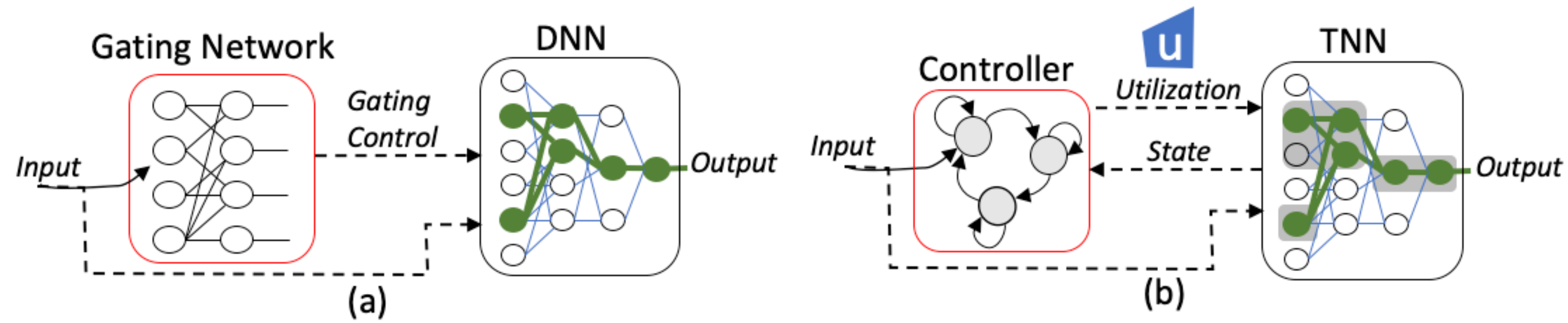}
\vspace{-0.5cm}
\caption{Conceptual architecture diagram. (a) Standard gating network, (b) Proposed dynamically throttleable network}
\label{fig:tnnarchitecture}
\end{figure*}

Our approach is fundamentally different from statically trained DNNs in that our goal is to reduce the \emph{working footprint} of inference, wherein only a subset of features needs to be computed for prediction. Figure \ref{fig:tnnarchitecture} shows the architectural diagrams of a standard gating network and our proposed TNN system. Several key elements for a TNN are described below.

First, TNN modules are filters structured into ``blocks'', while conditional gating is applied at that level with widthwise, depthwise, independent and nested strategies. The training approach for TNNs essentially enforces sparsity in the filters kernels such that there is no catastrophic accuracy loss as run-time dropout level increases in comparison to a naively trained network. TNNs are trained by maximizing task performance metric while minimizing the computational cost and within a single architecture for an application in conditions that can change over time. 

Second, TNNs use a \emph{single} input-dependent utilization parameter $u$ to conditionally gate throttleable modules. This design approach reduces the TNN control complexity while retaining highly flexible inference paths. We show the system with a data-driven controller that generates the \emph{utilization u} control signal based on the context learned from the input data. Then, the utilization and input are fed into the TNN for task-specific predictions. In comparison, standard gating networks are produces a large vector of control signals that makes it hard to train and tune the gating network.

Third, our TNN system uses a context-aware controller that decides which and how many TNN modules to ``turn off'' to obtain the best performance for a given level of utilization. A major benefit of this adaptive system is the separability of the controller and TNN. Any controller which provide a single scalar, namely the utilization parameter, can be used such as heuristic and learnable policies. That is, TNN controller can be as simple as a fixed utilization parameter, or a state-machine with policies trained via reinforcement learning.

We are contributing to the body of research on Conditional Computation and Adaptive Control (Section \ref{sec:RelatedWork}) with TNNs to improve model performance and address efficient inference with tight resource budgets such as memory, power, and computing horsepower. For vision-based systems, this is particularly important, as the workload and processing throughput are demanding and dynamic. To the best of our knowledge, we offer the following contributions in this paper:
\begin{itemize}
\item A novel architecture called throttleable neural networks with plug-and-play TNN modules comprised of convolutional and fully-connected layers controlled with a single utilization parameter.
\item A context-aware controller trained to regulate TNNs' performances and computing needs with learnable control policies.
\item Comprehensive experimental results on image classification, object detection and video classification tasks that demonstrate the effectiveness of TNNs.
\item A self-throttleable hand gesture recognition system which outperforms the vanilla architecture and all fixed utilization settings.
\item Physical power and run-time measurements on an embedded GPU showing the efficacy and practicality of TNNs.
\end{itemize}

This paper is organized as follows: in Section \ref{sec:RelatedWork}, we present an overview of related research. In Section \ref{sec:approach}, we briefly define the TNN system, including the architecture and training approaches using a controller. In Section \ref{sec:results}, we provide simulation results that evaluates our approach. We show a showcase example gesture recognition system built on the TNN framework. In Section \ref{sec:analysis}, we provide an analysis of the results with respect to the efficacy of TNN.  Finally, in Section \ref{sec:conclusions}, we conclude our work and contributions.

%% file: 2_RelatedWork.tex
\section{Related Work}\label{sec:RelatedWork}

We mainly focus on the research related to conditional computation and adaptive control of DNNs. Design of efficient architectures \cite{sandler2018mobilenetv2,tan2019efficientnet,zhang2018shufflenet} and Neural Architecture Search (NAS) \cite{45826,liu2018progressive,tan2019mnasnet,wu2019fbnet} are different from our approach such that we have multiple operation points within a single architecture for performance and efficiency trade-off rather than searching for a static model.  

\vspace{0.2cm}\noindent\textbf{Conditional Computation. } Our work builds on ``conditional computation'' where a small portion of the model is computed for prediction to reduce overall computational load. This work is related to sparse activations to turn off parts of the network. For example, CNMM \cite{Ruiz_2019_ICCV} is the ensemble of Convolutional Neural Networks (CNNs) providing intermediate classifiers; mixture-of-experts model \cite{shazeer2017outrageously} learns to rank DNN modules and selects only the top k modules to compute. Similarly, in \cite{teja2018hydranets}, a gating module is proposed to select among multiple branches of networks of features for each input. NestedNet \cite{kim2018nestednet} constructs a nested architecture with several levels of sparsity to meet resource requirements. Many approaches use ResNet modules \cite{he2016deep} as building blocks, leveraging the observation that ResNets behave as an ensemble of shallower networks \cite{veit2016residual}. Stochastic Depth \cite{huang2016deep}, Skipnet \cite{wang2018skipnet}, ConvNet-AIG \cite{veit2018convolutional} and BlockDrop \cite{wu2018blockdrop} are similar approaches that learn to bypass ResNet blocks based on the input. Similar ideas also appear in recent work on neural architecture search \cite{pham2018efficient}. Other examples include Adaptive Computation Time (ACT/SACT) \cite{figurnov2017spatially}, BranchyNet \cite{teerapittayanon2016branchynet}, and Dynamic Time Recurrent Visual Attention (DT-RAM) \cite{li2017dynamic}, where gating is based on early stopping once some threshold of ``confidence" is achieved. 

\vspace{0.2cm}\noindent\textbf{Adaptive Control. } More recently, the research is moving towards dynamically configuring the network topology at run-time based on input. In this line of research, no parallel experts are explicitly defined, and instead, a single network is computed by selectively activating model components such as units and layers for each input. For example, Slimmable Networks use switchable batch normalization for training and can adjust network width on the fly at run-time \cite{yu2018slimmable}. GaterNet uses a separate gating network to generate sparse binary controls for the backbone network in an input-dependent manner \cite{chen2019you}. Odena \etal \cite{odena2017changing} introduce a ``Composer" module to select the computational graph. D\textsuperscript{2}NN uses Q-learning to learn parameters of the control nodes that influences the computational path \cite{liu2018dynamic}. Similarly, Spasov \etal \cite{Spasov2019Dynamic} propose a channel based selection method by casting the gating function as a multi-armed bandit problem. Ahn \etal \cite{ahn2019deep} consider the dynamic network as an estimator-selector framework for multi-task learning such that each candidate network is optimized for one task.

\noindent\textbf{Summary. } The key difference for TNNs as compared to previous work is the flexible integration of the network modules with the control decision nodes. Compared with recent work theoretically, the TNN itself provides a much more flexible model selections and applications (Table \ref{tab:summary}). TNNs subsume the models of \cite{wu2018blockdrop,kim2018nestednet,yu2018slimmable}. In particular, \cite{yu2018slimmable,kim2018nestednet} are subsumed under our nested widthwise gating strategy with limited number of operational levels. \cite{wu2018blockdrop} is only applicable to ResNet-type networks with skip-connections which can be summarized as our nested depthwise gating strategy.

\vspace{-0.4cm} In terms of adaptive control, \cite{odena2017changing,chen2019you,liu2018dynamic,Spasov2019Dynamic} try to generate sequences of control nodes as a form of vectors. The control and backbone network are tightly coupled and joint before fine-tuning training is applied. Our approach on the other hand, introduces the single-scalar design of utilization parameter that is user-friendly and semantically meaningful. The advantages are that TNNs can preserve a monotonic performance without the controller while other work cannot guarantee this behavior. Furthermore, our two-phase training does not require fine-tuning the network jointly. 

\vspace{-0.4cm}Whereas most previous papers show only results for image classification, we also demonstrate our approach with insights and results for object detection and video classification tasks. Our approach leverages run-time filter selection inspired from Dropout \cite{Ba2013Adaptive,srivastava2014dropout} for TNN, and deep contextual reinforcement learning \cite{riquelme2018deep,liu2018dynamic,Spasov2019Dynamic} for the controller. Our control policies are learned to directly optimize application or user-defined performance metrics. 

\renewcommand{\arraystretch}{1.2}
\begin{table}[t]
\caption{Highlights of Conditional Gating Approaches }
\label{tab:summary}
\centering
\resizebox{0.9\columnwidth}{!}{%
\begin{tabular}{@{\extracolsep{12pt}}c| P{0.8cm} P{0.8cm} P{1.5cm} P{1.8cm} P{1.7cm} P{1.8cm} P{1.5cm}@{}}
\hline \hline
 & \multicolumn{3}{c}{Architecture Support} & \multicolumn{4}{c}{Reported Experimental Results}\\
 \cline{2-4}  \cline{5-8}
 &  Width-wise  & Depth-wise & Adaptive Control & Image classification & Object detection & Video classification & Embedded \\\hline
  \textbf{Proposed TNN} & \CheckmarkBold & \CheckmarkBold & \CheckmarkBold & \CheckmarkBold & \CheckmarkBold & \CheckmarkBold & \CheckmarkBold  \\
  BlockDrop \cite{wu2018blockdrop}  &  & \CheckmarkBold & & \CheckmarkBold &  &  &  \\
  NestedNet \cite{kim2018nestednet} & \CheckmarkBold\rlap{\scalebox{2}{$^{\ast}$}} & & & \CheckmarkBold  & \\
  DEN \cite{ahn2019deep} & \CheckmarkBold\rlap{\scalebox{2}{$^{\ast}$}} & \CheckmarkBold\rlap{\scalebox{2}{$^{\ast}$}} & & \CheckmarkBold  & \\
  GaterNet \cite{chen2019you} & \CheckmarkBold & \CheckmarkBold & & \CheckmarkBold &  \\
  D$^2$NN \cite{liu2018dynamic} & \CheckmarkBold & \CheckmarkBold & \CheckmarkBold\rlap{\scalebox{2}{$^{\ast}$}} & \CheckmarkBold &  \\
  Slimmable \cite{yu2018slimmable} & \CheckmarkBold &  & & \CheckmarkBold & \CheckmarkBold  \\
\hline
\multicolumn{4}{c}
\footnotesize{($\ast$ denotes limitations on number of operational levels).} \\
\end{tabular}
}
\end{table}

%% file: 3_TechApproach.tex
\section{Technical Approach}\label{sec:approach}

\subsection{Problem setting and objective}
We designed TNN with several properties that enable more flexibility for dynamic execution based on runtime context. Firstly, TNNs are composed of throttleable modules with task-specific gating modules that can be selected based on the input. Second, we coalesce gating control into a single utilization parameter to simplify the complexity of the control plane. Finally, a context-aware controller network is trained separately to directly optimize application-level performance metrics.

Our goal is to enable TNN with dynamic performance during inference while having same or similar number of parameters with vanilla ones. Furthermore, TNN should be modular such that (1) they are easily trained with gating policies that can be fixed or learned, (2) the controller can be trained separately without re-training the TNN, (3) the framework is largely model-agnostic.

\vspace{0.2cm}\noindent\textbf{TNN Definition. }
A neural network is a parameterized function $T_{\Phi}(x)$ mapping an input $x \in X$ to an output $\hat{y} \in Y$ where $x$ can be an image or video sequence, and $\hat{y}$ is the class prediction. We define a \emph{throttleable neural network} (TNN) as a function of two variables, $T_{\Phi}(x, u)$, where $u \in [0,1]$ is a control utilization parameter that indicates how much ``computational effort'' the network should exert. We emphasize that $u$ is an additional input to the network; after training is complete, the parameters $\Phi$ are fixed but $u$ can change. The general loss function of a TNN has two components,
\begin{equation} \label{eq:tnn-objective}
\mathcal{L}(x, u, y, \hat{y}) = L(y, \hat{y}) + \lambda C(x, u)
\end{equation}
where $y$ is the ground-truth label of $x$. The ``task loss'' $L$ is a task-specific performance measure, \eg, cross-entropy loss for classification. The ``complexity loss'' $C$ measures the resources used (FLOPs, energy, latency, \etc) for example $x$ at ``effort level'' $u$, and $\lambda$ controls the balance of the two losses.

\subsection{Throttleable Neural Networks}

\vspace{0.2cm}\noindent\textbf{Throttleable Module. } 
We consider the building block of TNN architectures that we call throttleable module (\emph{T-module}). A T-module consists of arbitrary number of filters within one layer such as convolutional and fully-connected layer or across several layers with skip-connections. Most CNN architectures can be converted into TNNs by replacing with T-modules. A T-module has the functional form
\begin{equation} \label{eq:gate_module}
t_\Phi(x, u) = \va(\vg_{\psi}(x, u) \odot \vf_{\Phi}(x))
\end{equation}
where $\vf_{\Phi}(x) = (f_1, \ldots, f_n)$ is a vector of \emph{components} conditioned on network parameters $\Phi$, $\vg_{\psi}(x, u) = (g_1, \ldots, g_n) : X \times [0,1] \mapsto \{0,1\}^n$ is the \emph{gating function} with parameters $\psi$, $\odot$ denotes element-wise multiplication, and $\va$ is the \emph{aggregation function} that maps $\vg_{\psi}(x, u) \odot \vf_{\Phi}(x)$ to the appropriate output space. For example of $g_i = 0$, the component $f_i$ is effectively disabled. The elements of $\vf$ can be arbitrary NN modules like convolutional filters or fully-connected layers, but we assume that they have the same input space and that their outputs can be aggregated appropriately like concatenation or summation. Our exposition focuses on a single gated module for simplicity, but in practice we compose multiple T-modules in a typical TNN such that $ y_{l+1} = t_\Phi(y_l, u)$ where the output of the $l-$th T-module is directly fed as the input of next T-module.

\begin{figure}[t!]
\centering
\begin{subfigure}[b]{0.20\linewidth}
        \includegraphics[width=\linewidth]{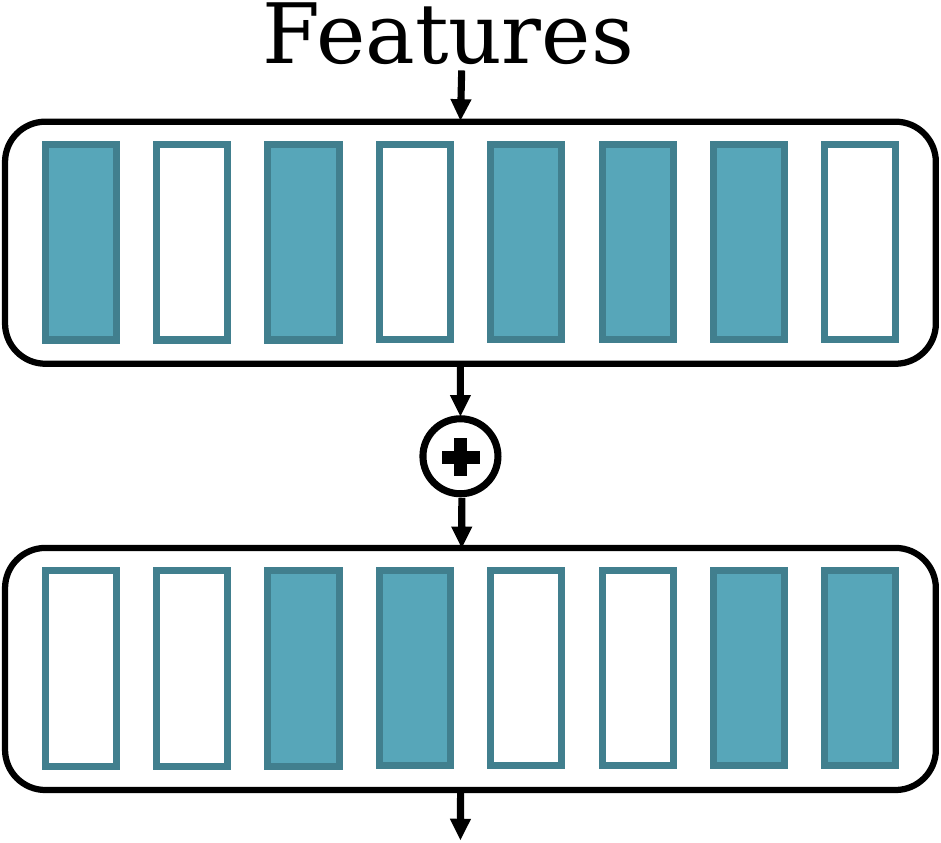}
        \caption{Widthwise}
        \label{fig:ww}
    \end{subfigure}
\begin{subfigure}[b]{0.20\linewidth}
        \includegraphics[width=\linewidth]{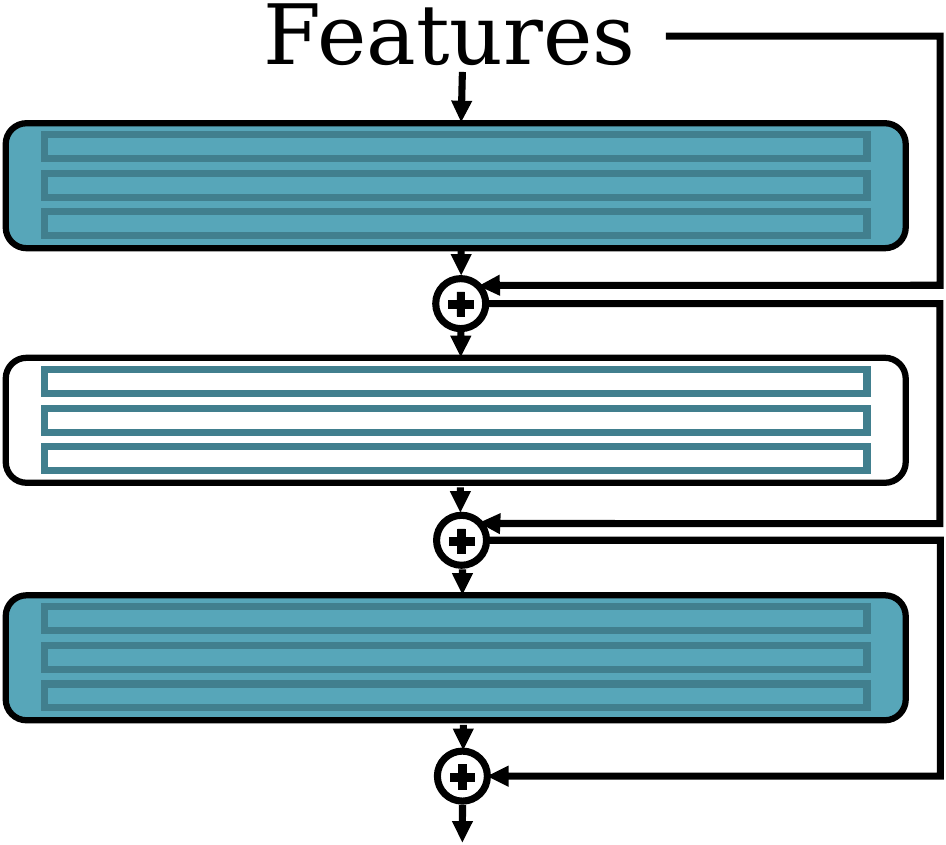}
        \caption{Depthwise}
        \label{fig:dw}
    \end{subfigure}
\begin{subfigure}[b]{0.20\linewidth}
        \includegraphics[width=\linewidth]{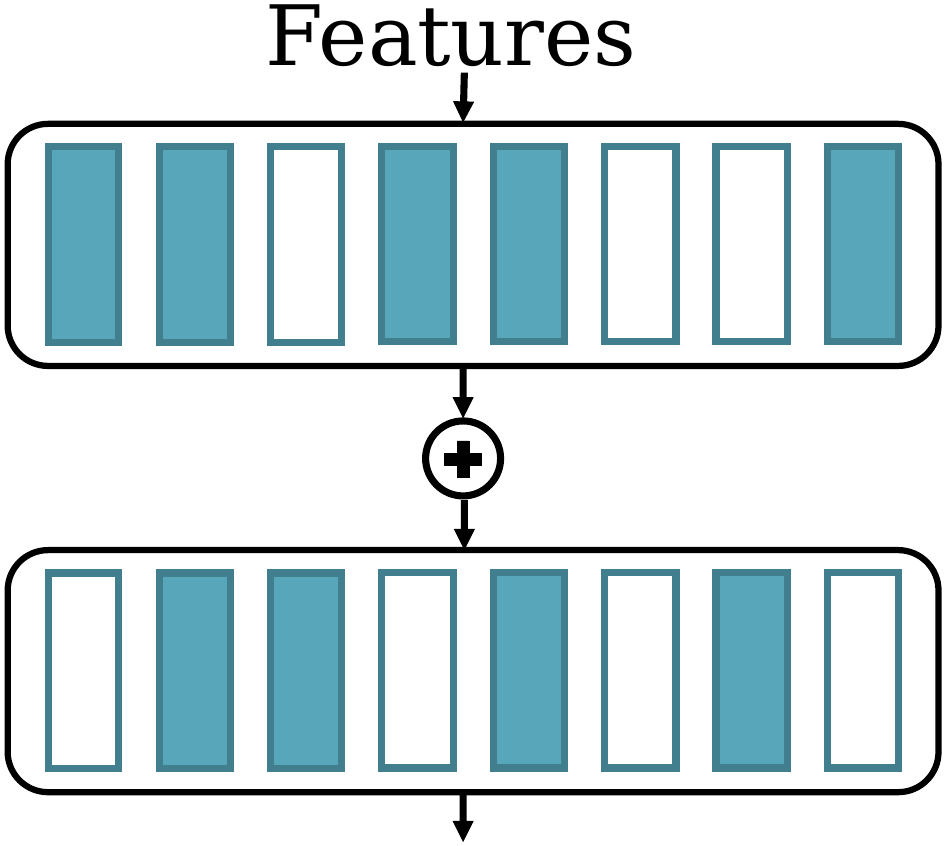}
        \caption{Independent}
        \label{fig:indep}
    \end{subfigure}
\begin{subfigure}[b]{0.20\linewidth}
        \includegraphics[width=\linewidth]{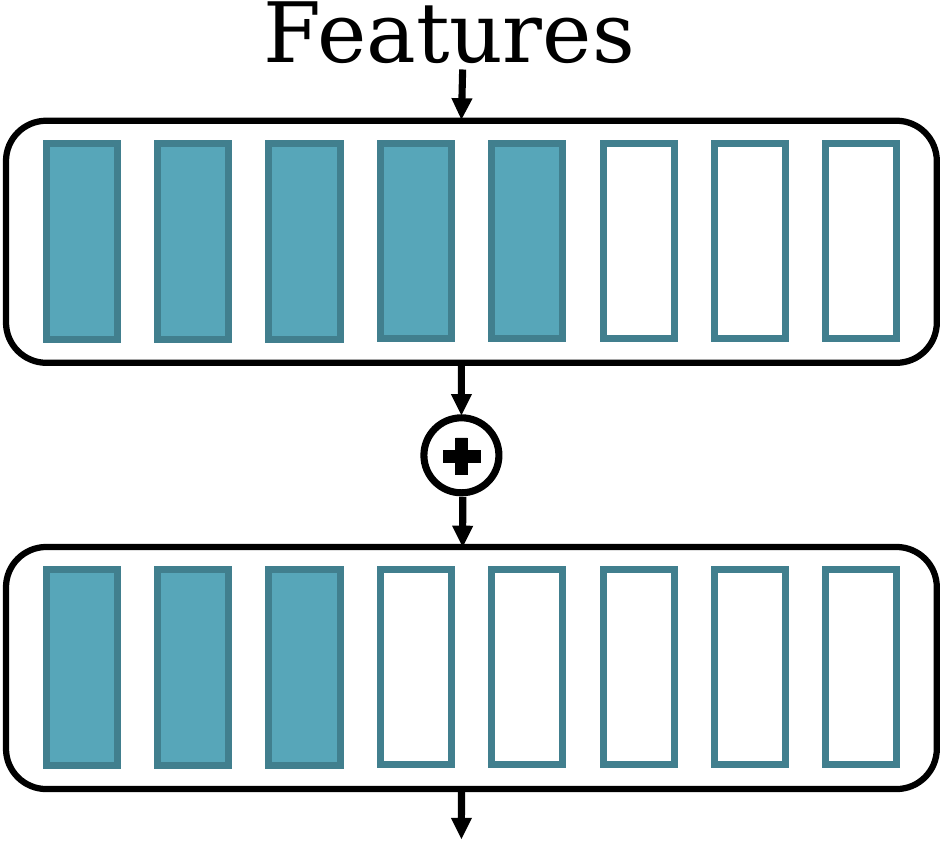}
        \caption{Nested}
        \label{fig:nested}
    \end{subfigure}
\caption{Selective gating strategies. The colored blocks are activated components while white blocks are gated. (a) and (b) are gating strategies over different dimensions. (c) and (d) have different ordering of gated components.}
\label{fig:gating}
\end{figure}

\vspace{0.2cm}\noindent\textbf{Gating Strategies.}
The components of $\vf$ could be anything from individual neurons to entire networks. We focus on an intermediate level of granularity. Decomposition into components can be defined along both the ``widthwise'' (number of features per layer in Figure \ref{fig:ww}) and ``depthwise'' (number of layers in Figure \ref{fig:dw}) dimensions. The widthwise and depthwise gating can also be used concurrently. Regarding the order of the gating, in previous conditional computation research, the components of each gated module are viewed as independent of one another with few constraints on their pattern of activation. This \emph{independent gating} scheme (Figure \ref{fig:indep}) works for each component to model different features, such as in \cite{shazeer2017outrageously,wu2018blockdrop}. For our goal of throttling over a range of set points, however, this specialization may produce some redundancy in the representation. We propose a different method that we call \emph{nested gating}. In the nested scheme, the gating function $\vg$ is constrained such that $g_i > 0 \Rightarrow g_j > 0 \; \forall j < i$ (Figure \ref{fig:nested}). Empirically, we observe that the nested scheme gives superior throttling performance given the same architecture.

\subsection{Context-Aware Controller}

Trained TNNs take an input and a single control parameter. However, using a fixed control parameter at run-time may not provide optimal trade-off between performance and utilization. The goal of incorporating the context-aware controller is to adjust the control parameter for TNNs dynamically, and there are many methods to achieve this functionality in a heuristic or learned manner. In this paper, we investigate a trainable input-dependent controller via solving a contextual bandit problem. Such a controller for the hand gesture recognition task is elaborated later, which we believe offers a generalizeable example for a wide variety of vision-based applications.

\vspace{0.2cm}\noindent\textbf{Input-dependent Contextual Bandit.} \label{sec:bandit}
A simplified approach to learn the controller is to frame it as a contextual bandit problem. Hypothetically, for a classification problem, the fact that different inputs or classes require different computation resources is demonstrated in \cite{wu2018blockdrop,chen2019you}. Based on this assumption, in order to determine what is the best utilization parameter for each input, the prediction of utilization parameter can be formulated as a sub-problem using contextual bandits.

\vspace{0.2cm}\noindent\textbf{State.} Given input image or video sequence $x$, the state representation is simply defined by the input itself as $s$. The goal of the agent is to receive current input signal and select actions in a way that maximises rewards for all similar input signals. It will learn appropriate utilization parameters $u$ for different states. The controller network will extract a dense feature representation of input, as well as potentially produce the probabilities for each action. The state reflects the contextual information associated with input $x$. 

\vspace{0.2cm}\noindent\textbf{Action.} The action is the selection of the utilization parameter $u$ from the action set $\mathcal{A}$ where $|\mathcal{A}| = K$. The actions are defined as arms in multi-armed contextual bandit problems. The learner pulls an arm according to information provided by input $x$. Therefore, the choice of actions is context-dependent. The number of actions $K$ relates to how $u$ is quantized. For the rest of the paper, we simply set $K=10$ such that $u \in \{ \frac{1}{k}\,|\, k = 1, \dots, K\}$. $K$ can be a arbitrary number but it should correlate with the number of components in T-modules. In our case, larger $K$ does not help since we have at most 16 components in each T-module.

\vspace{0.2cm}\noindent\textbf{Reward.} Given each input $x$, its label $y$, throttle ratio $tr_k$ of actual network computations over total full-throttled computations, and the prediction of TNN $\hat{y}$ with confidence $c$, the reward of performing action $a_k$ is defined as:
\begin{equation}
        r(a_k)=
        \left\{ \begin{array}{ll}
            \exp(1 - tr_k) * (1 - tr_k) & y = \hat{y} \\
            - (c + \gamma_1) * (tr_k + \gamma_2) & \text{otherwise}
        \end{array} \right.
\end{equation}
where $\gamma_1$ and $\gamma_2$ are constant parameters for reward balancing. In experiments, they are empirically set to $0.5$ and $1.5$ respectively. Such design of reward encourages to achieve higher accuracy while retaining low utilization. And more penalty is given if the prediction is wrong but takes a high utilization parameter.

\vspace{0.2cm}\noindent\textbf{Optimization.}
The contextual bandit neural network is trained via a policy gradient approach \cite{sutton2000policy}. The gradient of our objective follows the standard formulation:
$\nabla J(\Theta) = \mathbb{E}[\nabla\log \pi_{\Theta}(a|s)r(a)]$ where $\pi_{\Theta}$ is the parameterized neural network outputting the action per input. There are other designs of such a context-aware controller network. For example, we can utilize more complex interactions between the physical environment and the state of our system if we frame the controller as a full Markov Decision Process. In our formulation of the controller, we treat the TNN as a fixed model, but they can also be fine-tuned by end-to-end training.

\subsection{Training and Testing}

Training the TNN architecture is a two-phase procedure. During Phase-1 of TNN training, we train the feature representations of the TNN to be robust to varying amounts of gating. The throttling mechanism can be formulated as element-wise multiplication of a mask that ``gates'' a number of network intermediate features. Thus, the choice of how $u$ is sampled during training is important for obtaining the desired performance profile, which can follow different sampling methods of $u$. Naive training without gating can be viewed as one extreme, where we always set $u = 1$. Static gating functions map $u$ to a fixed subsets of the TNN while learnable ones adjust the TNN's paths on a per-input basis. 

There are several Phase-1 training strategies. One way is to sample utilization parameters randomly from uniform distribution, and the other is incremental training. For the latter strategy, the TNN is firstly trained with a low utilization parameter which only activates the first few components of each layer, then the number of active components is progressively increased. The second phase is to train a context-aware controller given the trained TNN with frozen parameters. During controller training period, the classification results and Multiply–Accumulate Operations (MACs) are used for computing the reward. The procedure of training a context-aware TNN is illustrated in Algorithm \ref{algo}.  

{\begin{center}
\scalebox{0.9}{
\begin{minipage}[b]{1.0\textwidth}
\vspace{-10pt}
\setlength\multicolsep{1pt}
 \begin{algorithm}[H]
 \caption{Context-aware TNN training}
 \begin{multicols}{2}
 \begin{algorithmic}[1]
 \renewcommand{\algorithmicrequire}{\textbf{Input:}}
 \renewcommand{\algorithmicensure}{\textbf{Output:}}
 \Require Training set $\{\mathbf{X}\}$
 \Ensure  Controller network $\pi_\Theta$, TNN $T_\Phi$ 

  \State Let $u \leftarrow{} u_0$, increment $\Delta u$, gating function $\vg$, aggregation function $\va$, learning rates $\alpha_1$, $\alpha_2$ 
  \For{epoch $\leftarrow{} 1, 2, \cdots, \mathcal{T}$} 
  \For{$\{(\mathbf{x}, \mathbf{y})\}$ from $\{\mathbf{X}\}$}
  \State $\hat{\mathbf{y}} \leftarrow{} $ Forward $ T_\Phi $ with Eqn \ref{eq:gate_module}
  \State $\Phi \leftarrow{} \Phi - \alpha_1 \nabla L_{cls}(\mathbf{y}, \hat{\mathbf{y}})$ 
  \EndFor
    \If {increase $u$}
  \State $u \leftarrow{} u + \Delta u$
  \EndIf
  \EndFor
  \State Freeze parameters of $T_\Phi$ 
  \Comment{Two-Phase TNN training}
  \For{epoch $\leftarrow{} 1, 2, \cdots, \mathcal{T}$}
  \For{$\{(\mathbf{x}, \mathbf{y})\}$ from $\{\mathbf{X}\}$}
  \State Forward $\pi_\Theta$, sample $(s, a)$ and $\mathbf{u} $ via $\epsilon-$greedy
  \State Forward $T_\Phi$, collect reward $r$ from $\hat{\mathbf{y}}$ and $tr_\mathbf{k}$
  \State $\Theta \leftarrow{} \Theta - \alpha_2 \nabla \log \pi_\Theta(a\, |\, s) r(a)$
  \EndFor
  \EndFor
 \State \textbf{return} $\pi_\Theta$, $T_\Phi$ 
 \end{algorithmic}
 \end{multicols}
 \label{algo}
 \end{algorithm}
 \end{minipage}%
 }
 \end{center}
 }
 
 Phase 2 training involves the controller learning the gating function while holding TNN parameters fixed. There are many methodologies to design controllers to manage the data network, such as heuristic rule-based approaches \cite{tann2016runtime}, data-driven model-based approaches using reinforcement learning (RL) \cite{liu2018dynamic}, \etc. Heuristic rule-based approaches are intuitive, hand-crafted, but not context-aware. Data-driven model-based approaches are capable of making context-aware decisions. For example, the controller is implemented as a much smaller neural network than the data network that takes in the input and predicts a proper control parameter. This is another advantage of a single controllable parameter as it is more feasible to learn.  

For learnable gating functions, phase-2 enforces the constraint that the actual complexity $c(\vg)$ should not exceed the target complexity $u$ by jointly optimizing the complexity term in the loss function (Eqn \ref{eq:tnn-objective}). The full objective $\mathcal{L}$ can be optimized via this two-phase training strategy.

%% file: 4_Experiments.tex
\section{Experiments}\label{sec:results}

\subsection{Experimental Setup}

To examine the generality of our TNN concept, we implemented gated convolutional and fully-connected layers as T-modules. Then, we created throttleable versions of several popular CNN architectures. All experiments are implemented with the same default architecture and training hyper-parameters for each dataset unless explicitly mentioned. We use suffix to indicate specific gating strategies such as ``-W'' for widthwise, ``-D'' for depthwise, ``-N'' for nested and ``-WN'' for widthwise nested, \etc. 

\vspace{0.2cm}
\noindent\textbf{VGG-W.} The VGG architecture \cite{Simonyan15} is a typical example of a ``single-path'' CNN. We apply widthwise gating with concatenation aggregation to groups of convolutional filters in each layer and combine the outputs by concatenating them forming a T-module. Because VGG lacks skip-connections, we enforce that at least one group must be active in each layer, to avoid making the output zero.

\vspace{0.2cm}
\noindent\textbf{ResNet-D.} We also experimented with a depthwise-gated version of standard ResNet with summation aggregation, similar to BlockDrop and SkipNet \cite{wang2018skipnet,wu2018blockdrop}. The T-modules are entire ResNet blocks that are skipped when gated off.

\vspace{0.2cm}
\noindent\textbf{ResNeXt-W.} ResNeXt \cite{xie2017aggregated} is a modification of ResNet \cite{he2016deep} that structures each ResNet layer into groups of convolutional filters that are combined by summation aggregation. We created a widthwise-gated version of ResNeXt (``ResNeXt-W'') by treating each filter group as a gated component.

\vspace{0.2cm}
\noindent\textbf{DenseNet-DW.} In the DenseNet architecture \cite{huang2017densely}, each dense block contains multiple narrow layers that are combined via concatenation. These narrow layers make natural units for gating. We view this architecture as both widthwise and depthwise gating due to the nature of dense blocks and skip connections.

\vspace{0.2cm}
 \noindent\textbf{C3D-W.} The T-module used in C3D is of the nested widthwise gating scheme. This gating strategy can be applied to 3D convolutional layers and wide fully-connected layers without skip connections, and widthwise gating is useful for grouping channels into fewer components.

\begin{figure*}[h]
\centering
\includegraphics[width=0.9\textwidth]{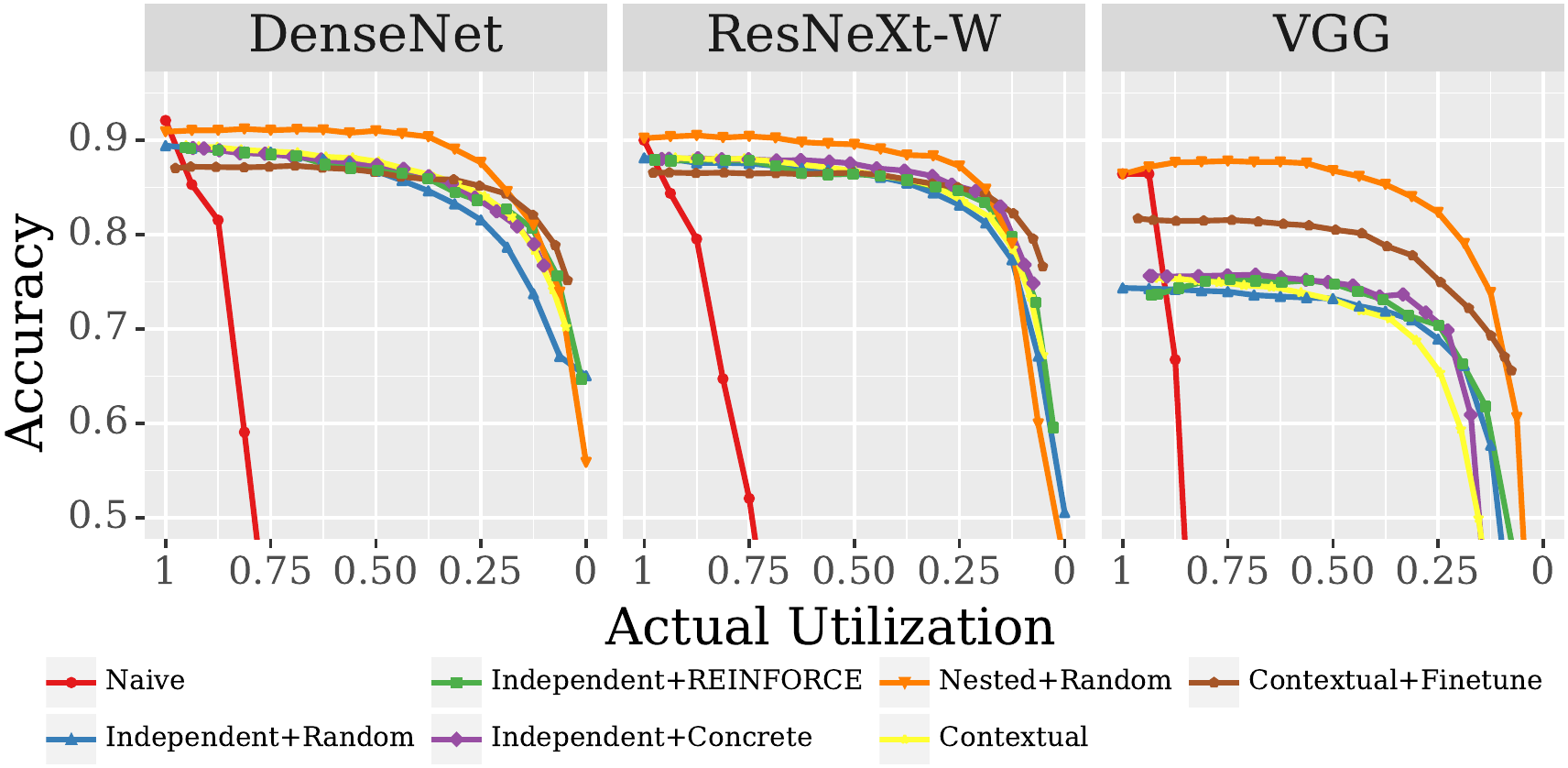}
\vspace{-0.3cm}
\caption{Comparison of classification accuracy with different gate control methods for three standard CNN architectures on the CIFAR-10 dataset.}
\label{fig:cifar10-blind}
\end{figure*}

\begin{figure*}[h]
\centering
\includegraphics[width=0.9\textwidth]{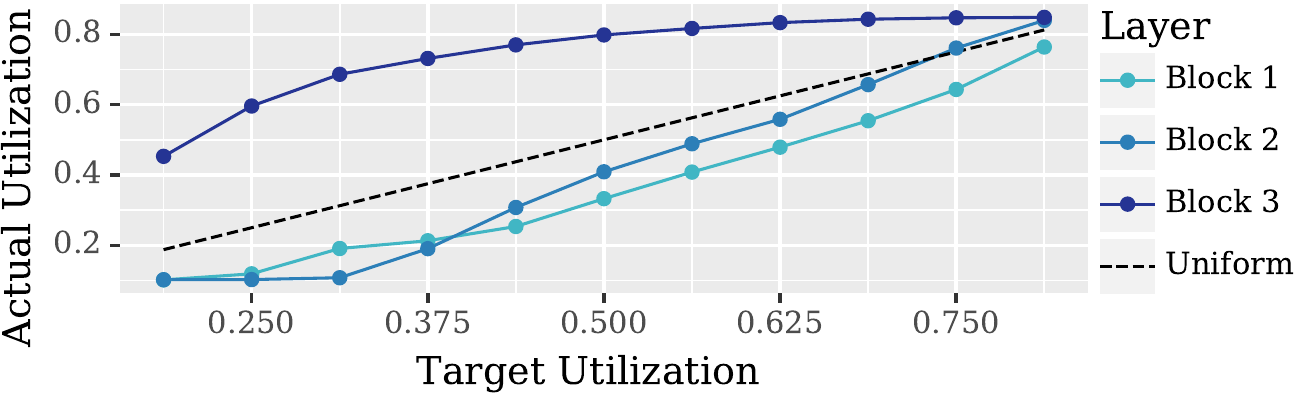}
\vspace{-0.3cm}
\caption{The learned pattern of component utilization for DenseNet-DW in CIFAR-10 with the REINFORCE training method. The dotted line shows uniform utilization.}
\label{fig:cifar10-utilization}
\end{figure*}


\subsection{Classification Task}
Our experiments compare approaches to creating TNNs using different gating strategies, with the goal of making a fair comparison between gating strategies using representative benchmark models and tasks. We evaluate the effectiveness of TNN and show that it has comparable peak performance and provides a family of sub-models given different setting of utilization. We also provide a comparison in Table~\ref{tab:comparetoothers}, showing TNN outperforming BlockDrop~\cite{wu2018blockdrop} by a large margin. However, we find that fair empirical comparisons with other approaches are difficult because of different concepts and network structures while our focus is the generalization of TNNs.

\vspace{-0.3cm}
\subsubsection{Image Classification on CIFAR-10.}
The CIFAR-10 dataset \cite{krizhevsky2009learning} is a standard image classification benchmark consisting of 32x32 pixel color images of 10 categories of object. We used the standard 50k image training set and 10k image test set, with no data augmentation. The CNN architectures are as follows. \textbf{DenseNet:} DenseNet-BC with 3 dense blocks having 16 components each with a growth rate of 12. \textbf{ResNeXt:} The ResNeXt architecture for CIFAR as described by \cite{xie2017aggregated} with 16 gated components in each of the 3 stages. \textbf{VGG:} The VGG-D architecture truncated to the first 3 convolution stages followed by a 4096 unit fully-connected layer; all three convolution stages and the fully-connected layer are partitioned into 16 gated components. The ``Independent+Learner'' methods use a ``blind'' control network (FC $\rightarrow$ ReLU $\rightarrow$ FC) that maps the control input $u$ to gate vectors $\vg$ for each gated module. We show results for the $C^2_{\text{dist}}$ complexity penalty (Eqn \ref{eq:tnn-objective}) where $C^p_{\text{dist}}(x, u) \Defined |c(\vg(x,u)) - u)|^p$ and $\lambda = 10$.

\vspace{0.2cm}
\noindent\textbf{Results.} The most noticeable result is that \emph{nested gating} substantially outperformed all variations on the independent method for all 3 architectures (Figure~\ref{fig:cifar10-blind}). The \texttt{Contextual} result is trained with a contextual controller and TNN independently, while \texttt{Contextual+fine} is trained jointly. The difference is especially pronounced for VGG; we attribute this to VGG learning more ``entangled'' representations than architectures with skip connections, which could make it more sensitive to exactly which components are gated off. For independent gating, the learned gating controllers are consistently better than random gating for both REINFORCE \cite{williams1992simple} and Concrete \cite{maddison2017concrete} training methods. The learned controllers achieve better performance by allocating computation non-uniformly across the different stages of the network shown in Figure~\ref{fig:cifar10-utilization}. Components in the later stages (higher-numbered blocks) are used preferentially over components in earlier stages. Note that the learned gating functions do not cover the entire range of possible utilization $[0,1]$. The useful range of $u$ is larger for larger $\lambda$ and for complexity penalties with $p = 1$.

\begin{figure}[t]
\centering
\begin{subfigure}[b]{0.49\linewidth}
        \includegraphics[width=\linewidth]{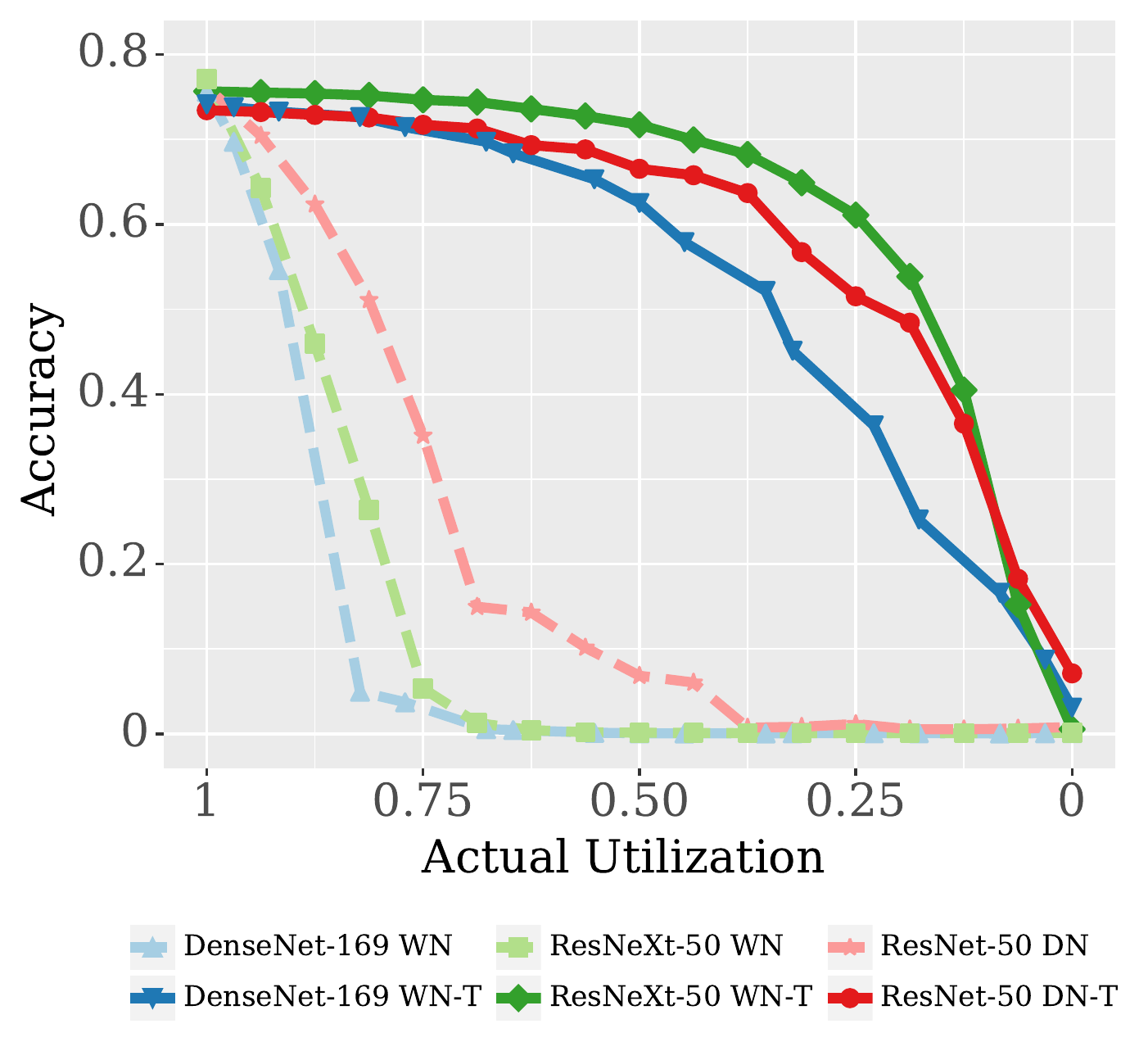}
        \caption{ImageNet}
        \label{fig:imagenet}
    \end{subfigure}
\begin{subfigure}[b]{0.49\linewidth}
        \includegraphics[width=\linewidth]{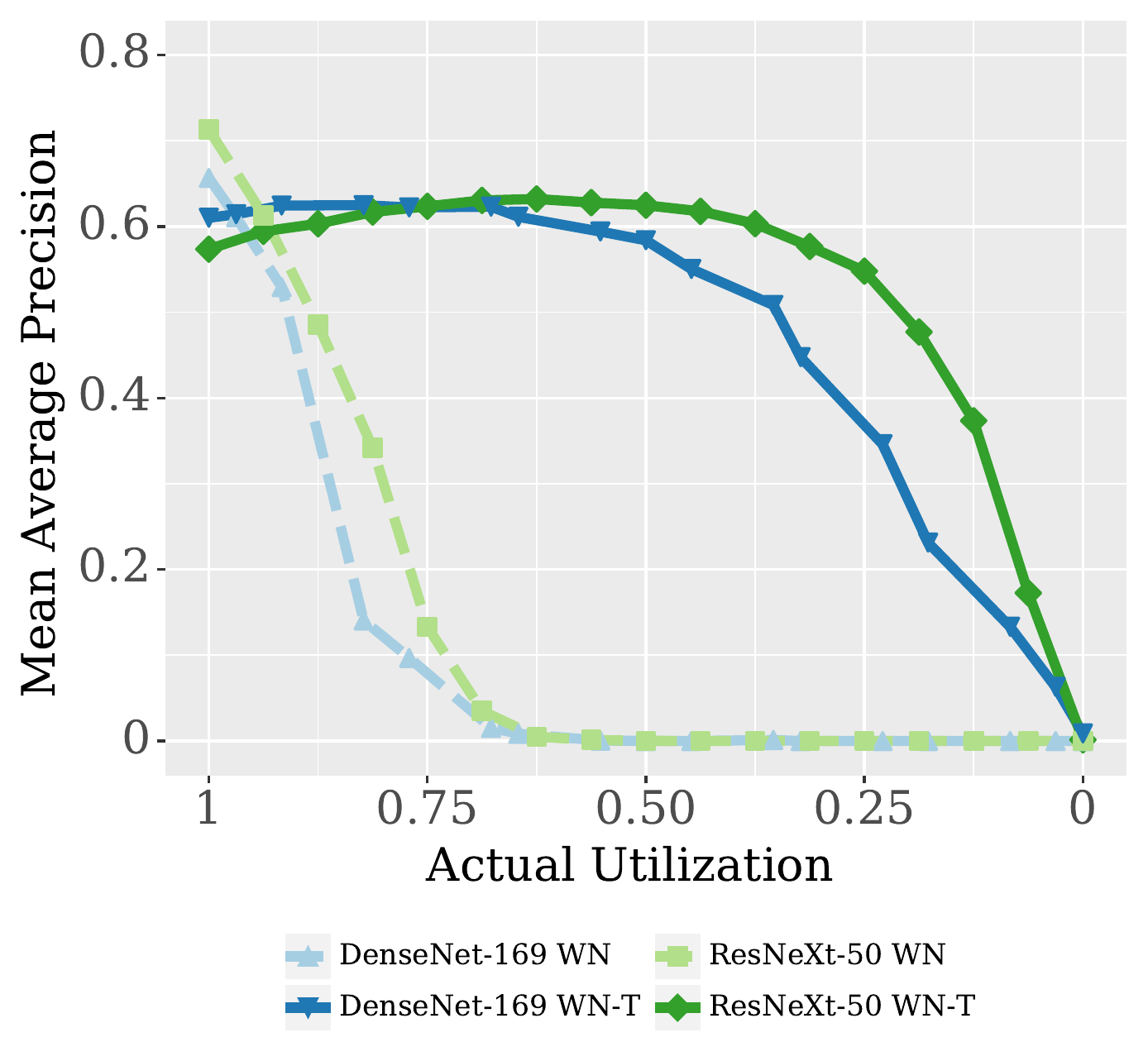}
        \caption{PASCAL VOC2007}
        \label{fig:voc2007}
    \end{subfigure}
\vspace{-0.2cm}
\caption{Comparison of throttleable architectures on larger-scale vision tasks. The ``-T'' suffix indicates \emph{throttling}. (a) Classification performance on ImageNet. (b) Object detection on VOC2007 using Faster-RCNN with a TNN ``backbone''.}
\label{fig:TNNCompare}
\end{figure}

\begin{table}[t]
\caption{Accuracy-Flops comparisons. MAC is converted to FLOP.}
\label{tab:comparetoothers}
\centering
\resizebox{0.7\columnwidth}{!}{%
\begin{tabular}{c|c|c|c}
\hline \hline
 
  Method  & Network Details & Accuracy & FLOPS \\\hline
  \multicolumn{4}{c}{CIFAR-10}\\\hline
   \textbf{Ours}  &  DenseNet-BC-3-Block N+R $u=0.375$ & \textbf{90.36} & \textbf{8.9M}\\\hline
   \textbf{Ours} & ResNeXt-W 3-stage $u=0.5$ & 89.55 & 15.9M\\\hline
   BlockDrop~\cite{wu2018blockdrop} & ResNet-32 without finetuning & 88.6 & $\approx$41.4M \\\hline
   \multicolumn{4}{c}{ImageNet}\\\hline
   \textbf{Ours} & ResNeXt-50-WN-T $u=0.875$ & \textbf{75.4} & \textbf{7.4G}\\\hline
   BlockDrop~\cite{wu2018blockdrop} & ResNet-101 & 75.2 & 9.9G\\\hline
\end{tabular}
}
\end{table}

\subsubsection{Image Classification on ImageNet.}

Our second set of experiments examines image classification on the larger-scale ImageNet dataset \cite{russakovsky2015imagenet} using the DenseNet-169, ResNeXt-50, and ResNet-50 architectures. For ImageNet, we used pre-trained weights to initialize the data path, then fine-tuned the weights with gating. In these experiments, we consider widthwise nested gating (``WN'' in Figure~\ref{fig:imagenet}) and depthwise nested gating (``DN''). In the DN scheme, we repeatedly iterate through the stages of the ResNet network from output to input and turn on one additional layer in each stage, unless the proportion of active layers in that stage exceeds $u$, and stop when the total utilization exceeds $u$. The ``-T'' suffix in the figures indicates that fine-tuning with gating is applied.

\vspace{0.2cm}
\noindent\textbf{Results.} The throttleable models reached a peak accuracy within about $2-3\%$ of the corresponding pre-trained model, and all are smoothly throttleable through the full range of utilization whereas the pre-trained models degrade rapidly with increased throttling. The ResNeXt model is the best in terms of both peak accuracy and area-under-curve.

\subsection{Object Detection} 

We next studied TNNs for the PASCAL-VOC2007 object detection task \cite{pascal-voc-2007}. To create a throttleable object detector, Faster RCNN \cite{ren2015faster} is selected and the ``backbone'' network is converted into a TNN. We use the DenseNet-169 and ResNeXt-50 in this experiment. Following the approach of \cite{he2016deep} for combining ResNet with Faster RCNN, we split our models after the layer with a $16\times16$ pixel receptive field, using the first half of the network as the feature representation, and the second half as the classifier. The vanilla models are trained on ImageNet and then fine-tuned on VOC2007. The throttleable versions take the throttleable networks from the ImageNet experiments and fine-tune on VOC2007 with gating.

We used the DenseNet-169 and ResNet-50 models from the \texttt{torchvision} package of PyTorch, and for ResNeXt-50 we converted the original Torch model \cite{xie2017aggregated} to PyTorch \cite{NIPS2019_9015} using a conversion utility \cite{clcarwin2017convert}. Our implementation of Faster RCNN is based on the open-source code of \cite{chen2017simple}. 

\vspace{0.2cm}
\noindent\textbf{Results.} Similar to results on image classification, we observe that the baseline method achieves higher peak mean average precision (MAP), but its performance quickly drops when any gating is applied (Figure \ref{fig:voc2007}). TNNs have little lower peak MAP, but degrade more gracefully with throttleable ability. During training, the $u$ is sampled uniformly. Consequently, the peak accuracy at $u=1$ is slight lower due to the compensations for the whole range of utilization parameters.

\subsection{Time Series} \label{sec:gesture}

\begin{figure*}[t]
\centering
\includegraphics[width=1.0\textwidth]{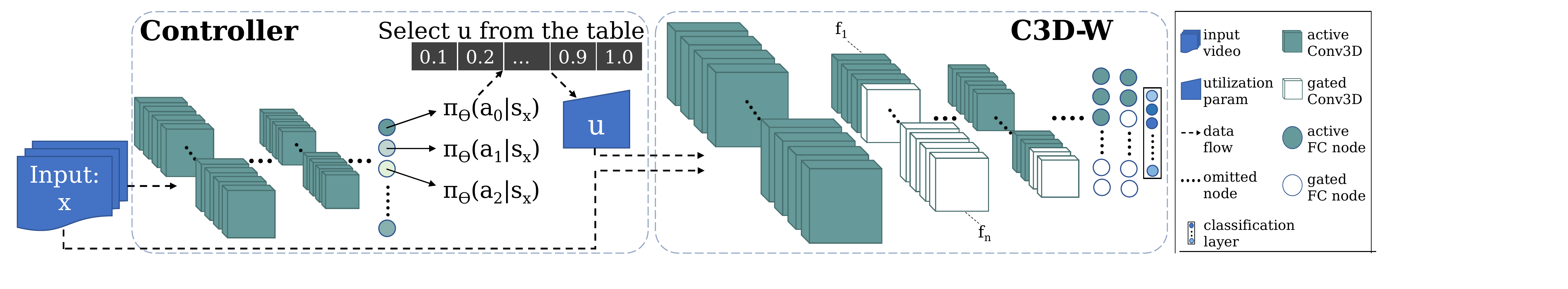}
\caption{A context-aware C3D-W framework for hand gesture recognition.}
\label{fig:architecture}
\end{figure*}

\vspace{0.2cm}\noindent\textbf{Hand Gesture Recognition.} \label{sec:gesture_system} To showcase the TNN architecture for more complex vision applications, we TNN architecture based on C3D \cite{tran2015learning} to perform hand gesture classification referred as C3D-W. It takes a fixed number of frames and a control parameter as input and outputs the gesture classification results. The controller is a deep contextual bandit network that follows the design discussed above. The framework is shown in Figure~\ref{fig:architecture}.

We implement a C3D-W architecture with widthwise nested gating strategy such that the first $n_{total} \times u$ components are activated. The controller network follows the approach in Section \ref{sec:bandit} and training strategy in Algorithm \ref{algo}. The deep contextual network is a small C3D network with only $1.9M$ parameters trained in a total of 20 epochs starting from $u=0.1$ with $0.1$ increment of $u$ in every two epochs. The optimizer RMSprop \cite{hinton2012neural} is used to train the context-aware controller. Experiments are performed for the hand gesture recognition task on the 20BN-JESTER dataset \cite{20bnjester}. There are in total of 148,092 videos of which 118,562, 14,787 and 14,743 are in training, validation and test set respectively. We report the test results on the validation set (not used for training). 

\vspace{0.2cm}\noindent\textbf{Training TNN}
The goal of training a throttleable network is to create a model that varies its complexity in response to the control parameter $u$. The natural measure of complexity is the number of active components, possibly weighted by some measure of resource consumption for each component,
\begin{equation}
c(\vg) = ||\vw||_1^{-1} \sum_i w_i \mathbf{1}(g_i \neq 0).
\end{equation}

The \emph{gate control strategy} embodied in $\vg_{\psi}(x, u)$ modulates the resource utilization of the TNN. Our experiments examine both static and learned gating functions. In the static approaches, the control parameter $u$ determines the number of gated blocks that should be used, and the choice of which blocks to turn on is made according to a fixed rule. Empirically, we find that a straightforward application of the \emph{nested} gating order works very well.

We also consider \emph{learning} the gating function. We enforce the constraint that the actual complexity $c(\vg)$ should not exceed the target complexity $u$ by optimizing the combined loss function $\mathcal{L}(x, u, y, \hat{y}) = L(y, \hat{y}) + \lambda C(x, u)$. We experimented with variants of $C$ of the two functional forms
\begin{align}
C^p_{\text{hinge}}(x, u) & \Defined \text{max}(0, c(\vg(x,u)) - u)^p \label{eq:hinge} \\
C^p_{\text{dist}}(x, u) & \Defined |c(\vg(x,u)) - u)|^p, \label{eq:dist}
\end{align}
for $p \in \{1,2\}$. In a sense, the ``hinge'' penalty (Eqn \ref{eq:hinge}) is the ``correct'' objective, since there is no reason to force the model to use more resources unless it improves the accuracy. In practice, we found that the ``distance'' penalty resulted in somewhat higher accuracy for the same resource use.

Learning the gate controller is complicated by the ``rich get richer'' interaction between $\vg$ and $\vf$, in which only the subset of $\vf$ selected by $\vg$ receives training, which improves its performance and reinforces the tendency of $\vg$ to select it. 

To address this, we adopt a two-phase training strategy similar to \cite{figurnov2017spatially}. In the first phase, we train the ``data path'' with random gating to optimize only $L$ while being ``compatible'' with gating. In the second phase, we train the gating controller to optimize the full objective $\mathcal{L}$ while keeping the data path fixed.

\vspace{0.2cm}\noindent\textbf{Training the Data Path.}
The data path is refered as the TNN for specific tasks without the controller. For classification problem, the data path is the network consisting of the feature extraction and classifier. For tasks except hand gesture recognition, the data network training follows the training schedule discussed here. During Phase~1 of training, we train the feature representations of the TNN to be robust to varying amounts of gating. The choice of how $u$ is sampled during training is important for obtaining the desired performance profile. From an empirical risk minimization perspective, we can interpret the training-time distribution of $u$ as a prior distribution on the values of $u$ that we expect at test-time. Ordinary training without gating can be viewed as one extreme, where we always set $u = 1$.

In our experiments, we employ a training scheme designed to maximize the useful range of $u$. For each training example, we draw $u \sim \text{Uniform}[0,1]$. Then, for each gated module, we select $k$ blocks to be gated on, where $k = \text{min}(n, \lfloor u \cdot (n+1) \rfloor)$ and $n$ is the number of blocks in the module. For \textsc{Nested} gating strategies, we set $g_1, \ldots, g_k$ to $1$ and $g_{k+1}, \ldots, g_n$ to $0$, while for \textsc{Independent} gating strategies we choose $k$ indices at random without replacement.

\vspace{0.2cm}\noindent\textbf{Training the Gating Module} Gating modules training is optional for some designs of gates. For \texttt{Contextual}, \texttt{Concrete} and \textsc{reinforce} gating schemes, this training is needed. When learning $\vg$, we proceed to Phase~2 of training, where we hold the data path parameters $\Phi$ fixed and optimize the gate controller parameters $\psi$. As in Phase~1, we draw the target utilization $u$ from a uniform distribution. We model the components of the gating function as Bernoulli random variables,
\begin{equation} \label{eq:gbern}
g_i(x, u; \psi) \sim \text{Bernoulli}(p_i(x, u; \psi)),
\end{equation}
and our task is to learn the function $\vp_{\psi}$ giving the activation probabilities of each component. Since $C$ is discontinuous, we need to employ a gradient estimator for training. We evaluated two existing methods of training networks with stochastic discrete neurons for this purpose.

\paragraph{Score function estimator} The most common approach \cite{bengio2013estimating,wang2018skipnet,wu2018blockdrop} is to treat $\vg$ as the output of a stochastic policy and train it with a policy gradient method such as the score function (\textsc{REINFORCE}) estimator,
\begin{equation*}
\nabla_{\psi} \E[J] = \E[J \cdot \nabla_{\psi} \log \text{Pr}(\vg_{\psi}(x, u))],
\end{equation*}
where $\text{Pr}(\vg_{\psi}(x, u))$ is the density of the random variable $\vg$. Since each $g_i$ is an independent Bernoulli random variable (Eqn \ref{eq:gbern}), the log probability is given by $\log \text{Pr}(\vg) = \sum_i \log [g_i p_i + (1 - g_i) (1 - p_i)]$.

\paragraph{Continuous relaxations} Relaxation approaches soften the discrete gate vector into a continuous vector of ``activation strengths.'' In particular, we use Concrete random variables \cite{maddison2017concrete} to stand in for discrete gating during training. Concrete distributions have a temperature parameter $t$ where the limit $t \rightarrow 0$ recovers a corresponding discrete distribution. The Bernoulli distribution is replaced by the binary Concrete distribution,
\begin{equation*}
\evg_i \sim \sigma((L + \log \alpha_i)^{-t}),
\end{equation*}
where $L \sim \text{Logistic}(0,1)$ and $\alpha_i = p_i/(1-p_i)$. We set $t > 0$ during training to make the network differentiable, and use $t = 0$ during testing to recover the desired hard-gated network.

\begin{table}[t]
	\small
	\begin{center}
	\resizebox{0.6\columnwidth}{!}{
		\begin{tabular}{l|l|l|l}
		\hline
		Stage & kernel dims & No. of co- & output dims \\
		  & $(d \times h \times w)$ & mponents & $(C \times D \times H \times W)$ \\ \hline
		conv-block-1 & $3\times 3 \times 3$ & 1 & $64 \times 16 \times 100 \times 160$ \\
		max-pool-1 & $1\times 2 \times 2$ & 1 & $64 \times 16 \times 50 \times 80$ \\
		conv-block-2 & $3\times 3 \times 3$ & 16 & $128 \times 16 \times 50 \times 80$ \\
		max-pool-2 & $2\times 2 \times 2$ & 1 & $128 \times 8 \times 25 \times 40$ \\
		conv-block-3 & $3\times 3 \times 3$ & 32 & $256 \times 8 \times 25 \times 40$ \\
		conv-block-4 & $3\times 3 \times 3$ & 32 & $256 \times 8 \times 25 \times 40$ \\
		max-pool-3 & $2\times 2 \times 2$ & 1 & $256 \times 4 \times 12 \times 20$ \\
		conv-block-5 & $3\times 3 \times 3$ & 32 & $512 \times 4 \times 12 \times 20$ \\
		conv-block-6 & $3\times 3 \times 3$ & 32 & $512 \times 4 \times 12 \times 20$ \\
		max-pool-4 & $2\times 2 \times 2$ & 1 & $512 \times 2 \times 6 \times 10$ \\
		conv-block-7 & $3\times 3 \times 3$ & 32 & $512 \times 2 \times 6 \times 10$ \\
		conv-block-8 & $3\times 3 \times 3$ & 32 & $512 \times 2 \times 6 \times 10$ \\
		max-pool-5 & $2\times 2 \times 2$ & 1 & $512 \times 1 \times 3 \times 5$ \\ \hline
		fc-1 & - & 16 & 512 \\
		fc-2 & - & 16 & 512 \\ 
		fc-class & - & 1 & 27 \\ \hline
		\end{tabular}
		}
	\end{center}
	\caption{The C3D-W architecture used in the paper for hand gesture recognition. If number of components is one, it simply means the layer is not a gated modular layer.}
	\label{tab:tnn_architecture}
\end{table}

\vspace{0.2cm}\noindent\textbf{C3D-W.} All of the object classification/recognition experiments are implemented using the PyTorch library (versions 0.3.1, 0.4, 1.0 and 1.3), and run on Nvidia GTX 1080 Ti and RTX 2080 Ti GPUs. We generate all of the performance curves by evaluating each model on the full test set using fixed values of \emph{u}. We refer to baseline models that do not have gating applied during training. Each data point in each chart is the result of a single evaluation run for a single instance of the trained model.

The input video dimension (number of frames, height, width) is $ [16, 100, 160] $. Table~\ref{tab:tnn_architecture} presents the exact specification of the C3D-W used for hand gesture recognition. The conv-block consists of a 3D convolutional layer with $ReLU$ activation. Stride of all convolutional layers is $1 \times 1 \times 1$. Stride of max-pool-1 is $1 \times 2 \times 2$, other max-pool layers are $2 \times 2 \times 2$. Padding of all conv-blocks is one, and padding of all max-pool layers is zero. The total number of parameters is $31.865M$ with a total of $48.703G$ MACs ($1.327G$ non-gated operations and $47.376G$ throttleable gated operations). 

For C3D-W training, we use four Nvidia RTX 2080 Ti graphic cards with the batch size of $20$ for each GPU. Adam optimizer \cite{kingma2015adam} is applied with an initial learning rate of $5\times 10^{-5}$, $\beta_1 = 0.5$ and $\beta_2 = 0.999$. The learning rate is reduced with patience of 3 epochs with default settings in PyTorch. The network is trained in 20 epochs (u is increased by 0.1 every two epochs), and no further fine-tune is applied. Note the training-time $u$ is generated incrementally which is different from the sampling method discussed earlier.

\vspace{0.2cm}\noindent\textbf{Context-Aware Controller} The controller network is a C3D architecture illustrated in Table \ref{tab:controller_specs}. Each conv-block consists of a convolutional, batchnorm and ReLU layer. Note that the controller network is not a TNN architecture. The total number parameters of the controller is $1.694M$ with a total of $34.077G$ MACs.
For controller training, we use a smaller batch size of 15 since it requires a GPU to handle both C3D-W and the controller network at the same time. We use RMSprop optimizer \cite{hinton2012neural} with a learning rate of $1 \times 10^{-7}$. The controller is trained for 10 epochs.

\begin{table}[t]
	\small
	\begin{center}
	\resizebox{0.6\columnwidth}{!}{
		\begin{tabular}{l|l|l}
		\hline
		Stage & kernel dims, stride, padding & output dims \\
		  & $(d \times h \times w)$  & $(C \times D \times H \times W)$ \\ \hline
		conv-3d-1 & $ 3\times 7 \times 7,\: 1 \times 1 \times 1,\: 1 $ & $ 64 \times 16 \times 96 \times 156 $ \\
		conv-3d-2 & $ 3\times 7 \times 7,\: 2 \times 2 \times 2,\: 1 $  & $ 64 \times 8 \times 46 \times 76 $ \\
		conv-3d-3 & $ 3\times 7 \times 7,\: 1 \times 1 \times 1,\: 0 $ & $ 64 \times 6 \times 40 \times 70 $ \\
		conv-3d-4 & $ 3\times 3 \times 3,\: 2 \times 2 \times 2,\: 1 $ & $ 64 \times 3 \times 20 \times 35 $ \\
		conv-3d-5 & $ 3\times 3 \times 3,\: 1 \times 2 \times 2,\: 1 $ & $ 32 \times 3 \times 10 \times 18 $ \\
		conv-3d-6 & $ 3\times 3 \times 3,\: 1 \times 2 \times 2,\: 1 $ & $ 32 \times 3 \times 5 \times 9 $ \\
		conv-3d-7 & $ 3\times 3 \times 3,\: 1 \times 2 \times 2,\: 1 $ & $ 16 \times 3 \times 3 \times 5 $ \\ \hline
		fc-1 & - & 256 \\
		fc-2 & - & 256 \\ 
		fc-action & - & 10 \\ \hline
		\end{tabular}
		}
	\end{center}
	\caption{The contextual controller architecture used in the paper for hand gesture recognition.}
	\label{tab:controller_specs}
\end{table}

\begin{figure}[h]
\begin{center}
\includegraphics[width=0.6\linewidth]{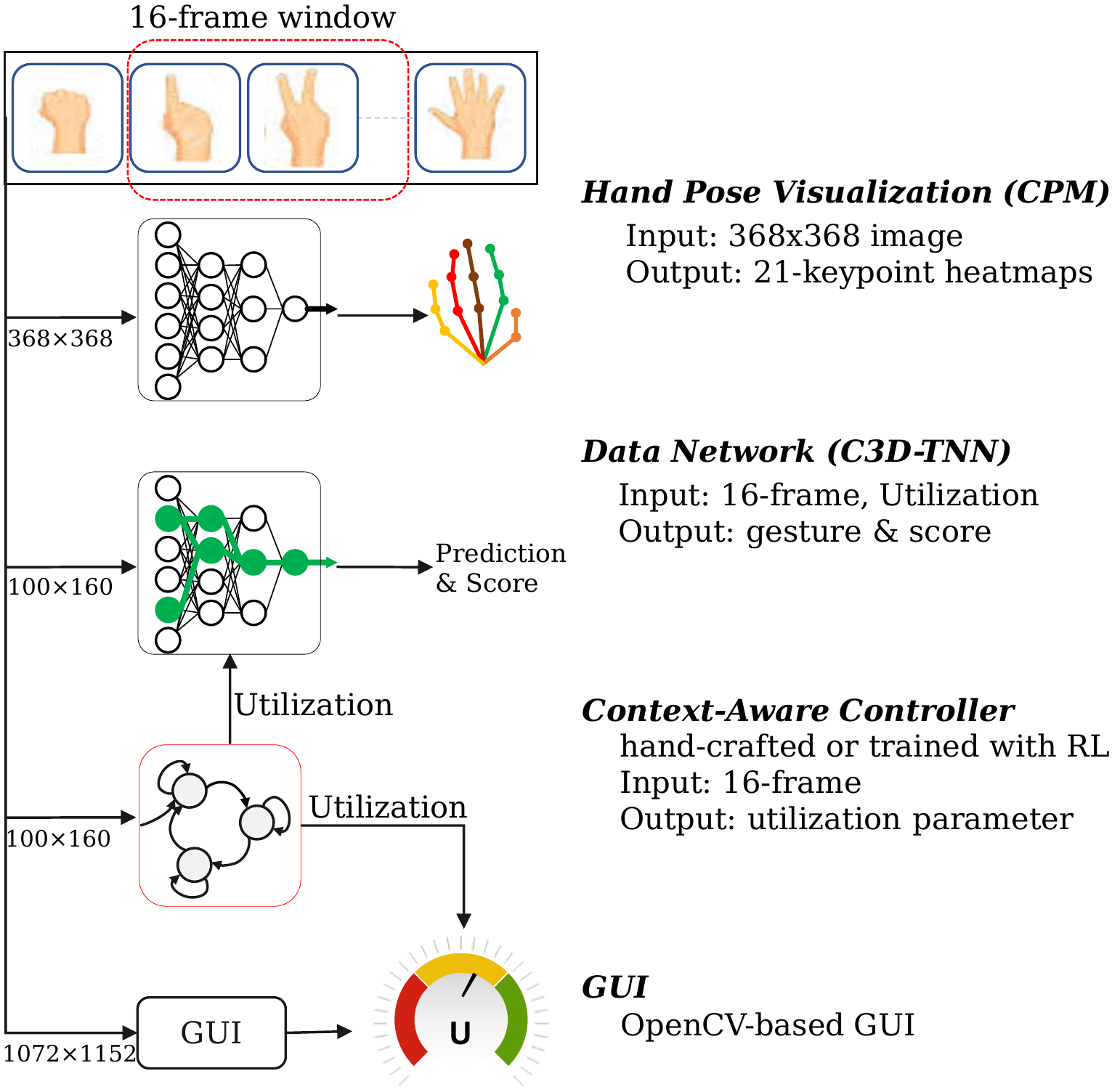}
\end{center}
\vspace{-0.4cm}
\caption{An application demo for hand gesture recognition.}
\label{fig:supp_demo}
\end{figure}

We have created a demo application to explain and showcase this work further. The demo comprises of the hand gesture recognition system (Figure \ref{fig:supp_demo}) that runs in real-time. The goals of the demo is to demonstrate the followings: (1) that spatio-temporal feature extraction can be performed in a dynamic manner on a TNN, and (2) the system can self-regulate its performance and computation needs. By casting the hand gesture recognition in real-time, we allow the controller to effectively decide on the utilization levels at the "onset" of the gesture. The decision making process is comprised of the control policy of the controller, which can be heuristically coded or trained. Computation needs can be adjusted during run-time while the whole model is preserved. 

In our demo, we have included a number of visualizations that better help the audience appreciate the inner workings of the demo. For example, we use Convolutional Pose Machines (CPM) \cite{wei2016convolutional} for estimating of 21 hand keypoints. It just provides visualizations in this application demo. Furthermore, we provide a \emph{Throttle Meter} in which we show the output of the controller (e.g. the single \emph{utilization} signal) so that the audience can directly see the self-regulating aspect of the system. We show the utilization signal as a GUI similar to a speedometer to make analogies to a gas-throttle in a vehicle (e.g. step on the gas to run harder). The interface is partially implemented using OpenCV \cite{opencv_library}. An overview video of the demo is available at \cite{Chai2020}.

\begin{figure}[h]
\centering
\includegraphics[width=1.0\linewidth]{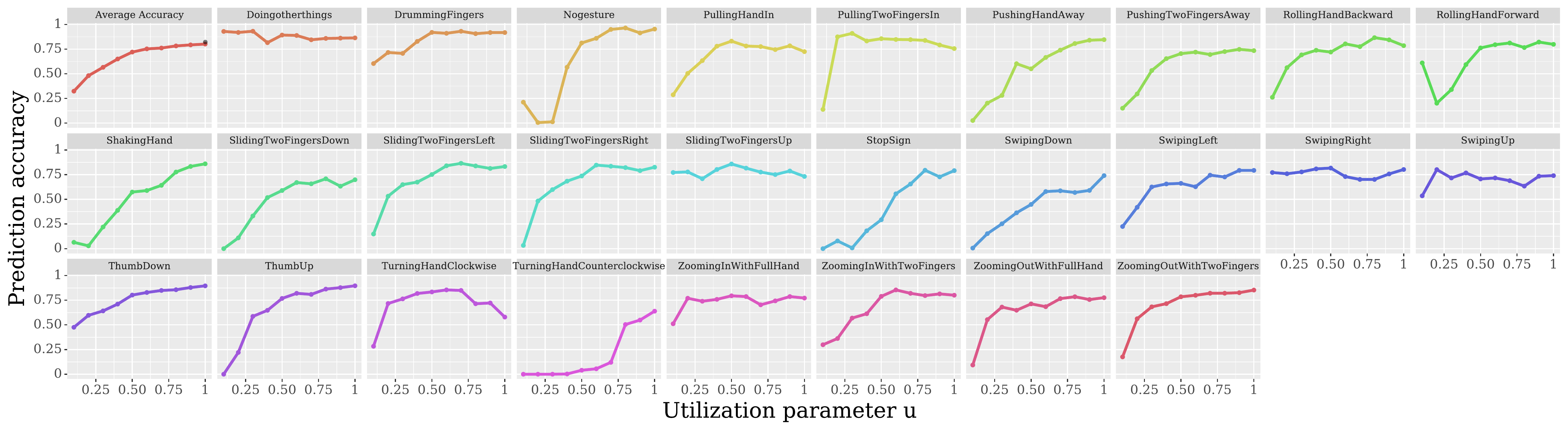}
\caption{Classification accuracy on validation set over utilization parameter $u$ for each gesture class. Top-left facet shows the average accuracy of all classes as well as the average accuracy of a vanilla C3D ($82.67\%$ denoted in black).}
\label{fig:c3d-tnn-results}
\end{figure}

\vspace{0.2cm}
\noindent\textbf{Results.} Figure \ref{fig:c3d-tnn-results} shows the various utilization levels for each gesture class. We verify our earlier hypothesis (Section \ref{sec:bandit}) that some classes require more resources than others. For class `\texttt{Doingotherthings}', the classification performance remains almost constant for each tested utilization parameter $u$ while it is linear with $u$ for the class `\texttt{Shakinghand}'. Given this trained TNN, the classification results of each test video for every $u$ parameter are collected. Then a performance upper bound on the total utilization with respect to the controller is derived. The expected predictions of a trained controller should be close to the upper bound with $96.14\%$ accuracy and total utilization of 4050.5 (sum of all the utilization parameter of each correctly-classified test data). Note that the upper bound is an ideal case for a particular TNN.

\begin{figure}[h]
\begin{center}
\includegraphics[width=1.0\linewidth]{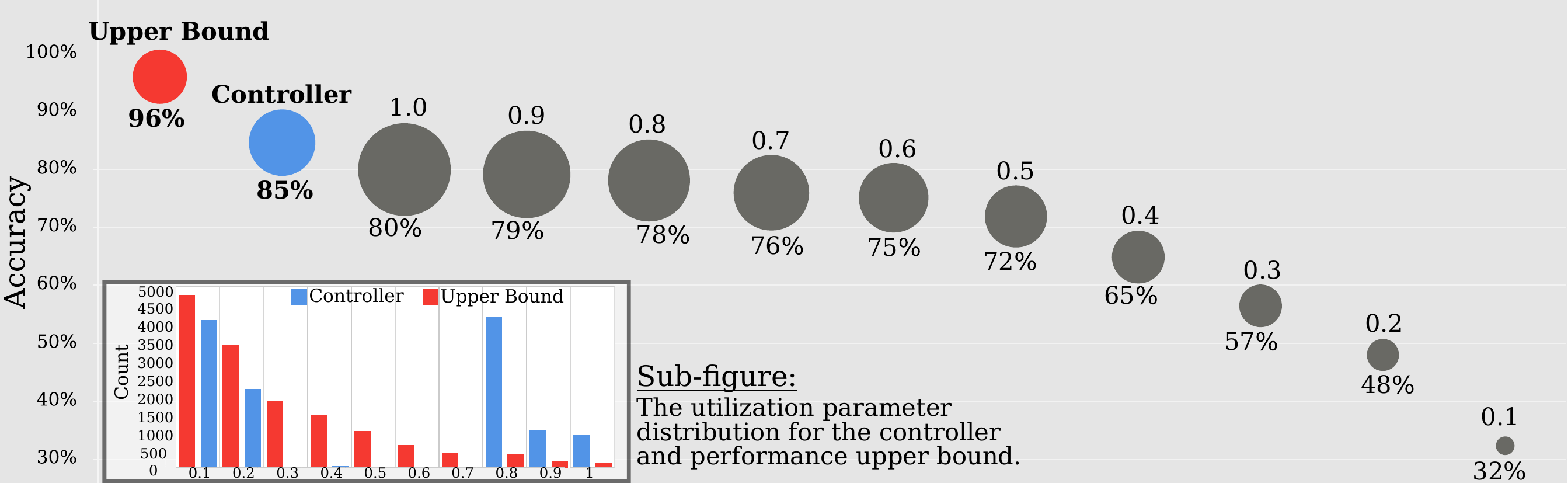}
\end{center}
\caption{Results of accuracy over total utilization and utilization distributions on validation set. The area of circle denotes the total utilization. The controller achieves the best accuracy-computation trade-off.}
\label{fig:controller}
\end{figure}

The effectiveness of this context-aware controller is confirmed with Figure \ref{fig:controller}. The learned controller achieves $85.24\%$ accuracy while vanilla C3D achieves $82.67\%$. Recall that the C3D-W has little performance drop compared with vanilla C3D at $u=1$. It manifests that the controller learns the input-specific utilization for each input. In the sub-figure of Figure \ref{fig:controller}, by comparing the utilization distributions, the controller learns a sparse selection of $u$ where $u=0.1$, $u=0.2$ and $u=0.8$ are chosen more often. Although the controller has $13.47\%$ more utilization and $10.90\%$ accuracy drop compared to the upper bound, it outperforms all the fixed utilization settings as well as the vanilla C3D. It further demonstrates how the system can operate within different utilization levels with a learned control policies. Regarding the time of processing a 16-frame video sequence, the inference time for the vanilla C3D, C3D-W and the controller is 74.7 ms, 97.3 ms (on average) and 12.1 ms respectively. The controller adds a little computation cost but reduces much more overall computations for TNNs.

%% file: 5_Analysis.tex
\section{Analysis}\label{sec:analysis}


TNN enables a framework for conditional computation for DNNs, whereby the overall computational load and model accuracy can be determined dynamically at inference time. In this section, we offer in depth analysis of our results to best evaluate the proposed TNN architecture and methodology. We start with embedded results to show practicality of implementations on hardware platforms. We evaluate the proposed formulation and end-to-end training for TNN towards its overall efficiency and benefits.

\subsection{Hardware Implementation}
We examine the performance of TNN on embedded systems. We implement and measure the computational benefits of TNNs on a NVIDIA Jetson AGX Xavier processor \cite{Nvidia2019Jetson}, using a throttleable VGG-D model trained on the CIFAR-10 dataset. Relative savings in run-time and power rather than actual peak values are shown because the results depend highly on provided hardware mechanisms. We replace the convolution operations in VGG with tensor slicing described in Algorithm~\ref{alg:slice-gating} where $W$ and $b$ are the weights and bias for the convolution kernel. To perform a gated convolution, we form truncated weights and bias tensors by removing channels corresponding to gated-off components, perform an ordinary convolution with the truncated weights, and restore the output to its original shape by padding zeros. In the case of nested gating, the implementation is highly efficient, requiring only two tensor slice operations and one concatenation.
{\centering
\begin{figure}[h]
\scalebox{1.0}{
\begin{minipage}[b]{0.4\textwidth}
\begin{algorithm}[H]
\begin{algorithmic}
	\State{NestedConv2D-T}{($x$, $u$, $W$, $b$)}
		\State Let $C = \text{channels}(W)$
		\State Let $n = \lceil u \cdot C \rceil$
		\State $\hat{W}, \hat{b} \leftarrow W[0:n], b[0:n]$
		\State $\hat{y} \leftarrow$ \Call{Conv2D}{$x$, $\hat{W}$, $\hat{b}$}
		\State $z \leftarrow$ \Call{Zeros}{$n - C$, $\text{size}(\hat{y})$}
		\State $y \leftarrow$ \Call{Concatenate}{$\hat{y}$, $z$}
		\State \Return $y$
\end{algorithmic}
\caption{Conv2D-N}
\label{alg:slice-gating}
\end{algorithm}
\end{minipage}
}
\hfill
\begin{minipage}[b]{0.60\textwidth}
\begin{center}
\includegraphics[width=1.0\linewidth]{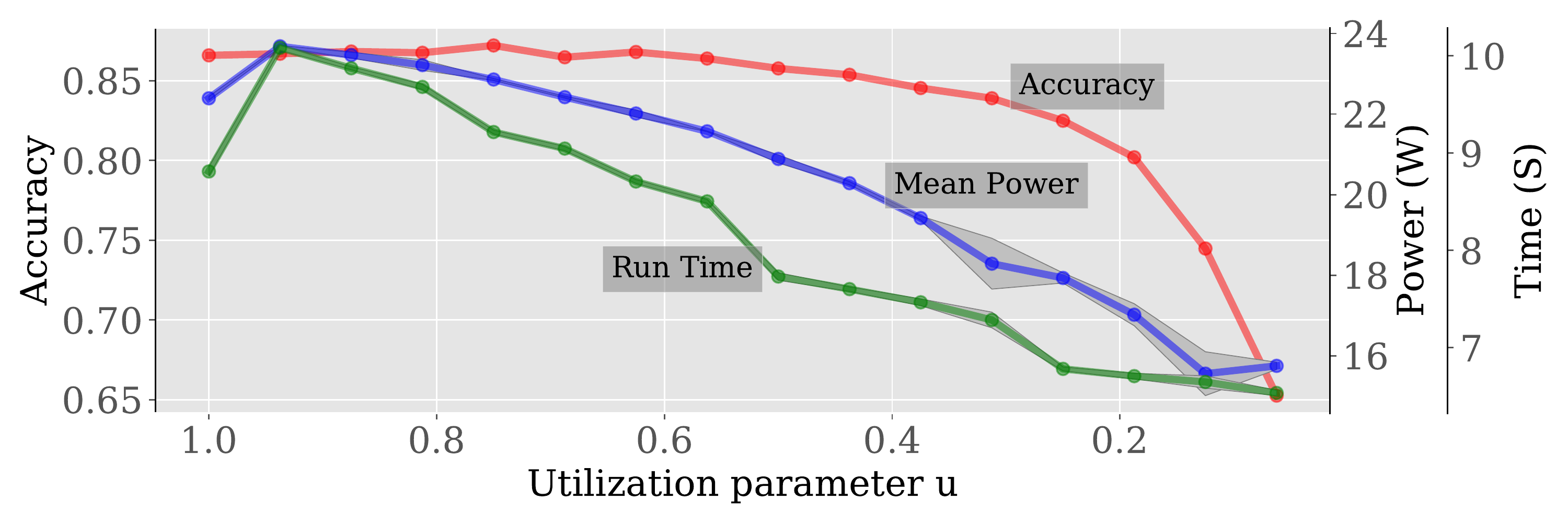}
\end{center}
\caption{Measured throttling performance on NVIDIA Jetson AGX Xavier.}
\label{fig:xavier}
\end{minipage}
\end{figure}
}

Figure~\ref{fig:xavier} shows accuracy, mean power draw, and run time for classifying the entire CIFAR-10 test set on the Xavier GPU in ``MAXN'' power mode. The shaded regions show standard deviation of 3 experiments. Salient aspects of this result is the linear ramp-down of the power and run-time while accuracy remains high at lower utilization settings. Compared to no throttling, the minimum throttle setting tested used $70\%$ of the power, $74\%$ of the run time, and $52\%$ of the total energy. We anticipate similar gains in TNN inference efficiencies for other TNN models described in this paper. The key is having an effective control-policy to guide the utilization settings to match application needs.

\subsection{Architecture}
\vspace{0.2cm} 
\noindent\textbf{Conditional Gating.} Our TNN formulation is based on a \emph{uniform} control plane using a single generalized TNN control parameter (e.g. utilization \emph{u}). Using the single \emph{u} control parameter, we are able to simplify the TNN modules and generalize the gating strategies (widthwise, depthwise, independent, and nested). Typically, there are lots of parameters in a DNN, and so, existing approaches are focused on the multitude of control signals to ''gate'' a DNN. With TNN, the granularity of the control is coarser grain compared to other approaches using finer grain (but more complex) signals.

Using uniform TNN modules helps decouple the end-to-end training of the model into two-phases. This allow us to simplify the learning objectives into different and separate levels of concerns in each phase. The TNN modules learn features that are robust of runtime dropout by anticipating that its contribution to the overall task may be gated off. Meanwhile, the controller learns a control strategy to maximize the reward (e.g. which modules to gate off to minimize computational load for that control signal \emph{u}).  In comparison, other conditional gating approaches require joint training with both gating network and the DNN, resulting in a training scheme that is more complex and may affect model convergence (e.g. \cite{wang2018skipnet}, \cite{yu2018slimmable}, and \cite{chen2019you} to name a few).

This two-phase training approach also means that TNN modules learn differently than other conditional gating approaches.  Standard conditional approaches overlay a gating structure on top of a network trained for strictly for accuracy. In comparison, TNN module learns to anticipate the utilization \emph{u} parameter, and thus will select weights that produces the best overall result for any value of \emph{u}. This is the main difference that decouples the controller in TNN. Our results show that such the TNN gating approach enables a more flexible throttling performance with model convergence, even with a single utilization \emph{u} parameter.

\vspace{0.2cm} 
\noindent\textbf{Cost of Inference.} We designed the TNN framework to have low overhead in conditional gating. We show that TNN can enable dynamic inference peformance with the same or similar number of model parameters. The abstraction afforded by the utilization \emph{u} parameter enables us to train TNN without additional parameters, and instead, we train TNN with runtime dropout levels based on different \emph{u} levels.

Training with runtime dropout allows the TNN modules to be regularized based on a more robust distribution, akin to an attention mechanism that selects only subset of the network for a given task. With TNN, the gating overhead is low because we build in this overriding prior where any TNN module can be gated off. A naive approach would just add a control plane with mask or gating signals specifically designed for that specific network. Such a control plane adds complexity and additional training parameters that we circumvented with the TNN framework.

Having a uniform control signal \emph{u} also allows us enable adaptive throttling to other inputs beyond those in the input data set. For example, a deployed application may run differently based on environmental conditions (such as battery charge, illumination levels, etc.), and as such, may require alternative operating conditions for TNN. Because we have decoupled the TNN module from the controller, we can also throttle TNN based on application inputs. For example, a system with low battery charge may impose a lower utilization threshold using the \emph{u} utilization parameter, and thereby dynamically adjusting the quality of services based on system capability.

\vspace{0.2cm}
\noindent\textbf{Controller Efficacy.}
There are many methodologies to design controllers to manage the data network (TNN), such as heuristic rule-based approaches \cite{tann2016runtime}, data-driven model-based approaches using reinforcement learning (RL) \cite{liu2018dynamic}, \etc. Heuristic rule-based approaches are intuitive, hand-crafted, but not necessarily context-aware. Data-driven model-based approaches are capable of making context-aware decisions, and as shown in this paper, reinforcement learning (RL) methods can be used to learn policies that manipulate the utilization parameter \emph{u} directly. 

TNNs work well when predictions and decisions in a dynamic environment are hard to make, \eg, in video-related tasks where temporal action and contextual information in the video sequences is changing dynamically. With the context-aware controller, the system is directly optimized for application level performance metrics, and not just weight sparsity based on the training dataset. As shown in our C3D architecture for gesture analysis, our control policies can be trained using Contextual Bandits \cite{langford2007epoch} or Markov Decision Processes \cite{puterman2014markov} aimed at predicting future states for the TNN. This enables TNNs to seamlessly throttle through the appropriate utilization levels based on dynamic, global level inputs.

In our TNN framework, context-aware controllers are decoupled from the main network. With an RL-based approach, rewards can be used to learn policies that determine the decision on how much to throttle and which components to turn off. It is also possible to train different controllers after deployment to enable different policies or additional system/application inputs. In our experience, the controller is implemented as a much smaller neural network than the data network that takes in the input and predicts a proper control parameter. This is another advantage of a single controllable parameter as it is easier to train.

%% file: 6_Conclusion.tex
\section{Conclusions}\label{sec:conclusions}
In this paper, we presented a novel run-time \emph{throttleable} neural network, as an adaptive model with flexible topology whose performance can be varied dynamically to produce a range of trade-offs between task performance and resource consumption. We designed TNNs using T-modules that can be activated and deactivated during inference time. A separately trained context-aware controller which is capable of input-dependent resource management was implemented. Comprehensive results on image classification and object detection show that TNNs can be effectively throttled across a range of set points, while having peak accuracy comparable to their vanilla architectures. The experimental results on hand gesture recognition task demonstrate that the dynamic execution of TNNs with a context-aware controller outperforms the vanilla architecture and all fixed settings of utilization parameter.